\documentclass[11pt,a4paper]{article}
\usepackage[hyperref]{acl2019}
\usepackage{times}
\usepackage{latexsym}
\usepackage[normalem]{ulem}

\usepackage{multirow}
\usepackage{arydshln}
\usepackage{amsmath}

\usepackage{subfigure}

\usepackage{url}

\usepackage{graphicx}
\aclfinalcopy 
\def\aclpaperid{***} 

\newcommand{\dd}{Domain Differential Adaptation}
\newcommand{\ddab}{DDA}

\title{{\dd} for Neural Machine Translation}

\author{Zi-Yi Dou, Xinyi Wang, Junjie Hu, Graham Neubig \\
  Language Technologies Institute, Carnegie Mellon University \\
  {\tt \{zdou, xinyiw1, junjieh, gneubig\}@cs.cmu.edu}}

\date{}

\begin{document}

\maketitle
\begin{abstract}
  Neural networks are known to be data hungry and domain sensitive, but it is nearly impossible to obtain large quantities of labeled data for every domain we are interested in.
  This necessitates the use of domain adaptation strategies. 
  One common strategy encourages generalization by aligning the global distribution statistics between source and target domains, but one drawback is that the statistics of different domains or tasks are inherently divergent, and smoothing over these differences can lead to sub-optimal performance. 
  In this paper, we propose the framework of {\it Domain Differential Adaptation (DDA)}, where instead of smoothing over these differences we embrace them, directly modeling the difference between domains using models in a related task.
  We then use these learned domain differentials to adapt models for the target task accordingly. Experimental results on domain adaptation for neural machine translation demonstrate the effectiveness of this strategy, achieving consistent improvements over other alternative adaptation strategies in multiple experimental settings.%
\footnote{Code is available at \url{https://github.com/zdou0830/DDA}.}
\end{abstract}



\section{Introduction}
\label{sec:introduction}



Most recent success of deep neural networks rely on the availability of high quality and labeled training data \cite{he2017mask,Vaswani:2017:NIPS,povey2018semi,devlin2018bert}. 
In particular, neural machine translation~(NMT) models tend to perform poorly if they are not trained with enough parallel data from the test domain~\cite{koehn2017six}.
However, it is not realistic to collect large amounts of parallel data in all possible domains due to the high cost of data collection.
Moreover, certain domains by nature have far less data than others.
For example, there is much more news produced and publicly available than more sensitive medical records.
Therefore, it is essential to explore effective methods for utilizing out-of-domain data to train models that generalize well to in-domain data.

\begin{figure}[t]
\centering
\includegraphics[width=0.49\textwidth]{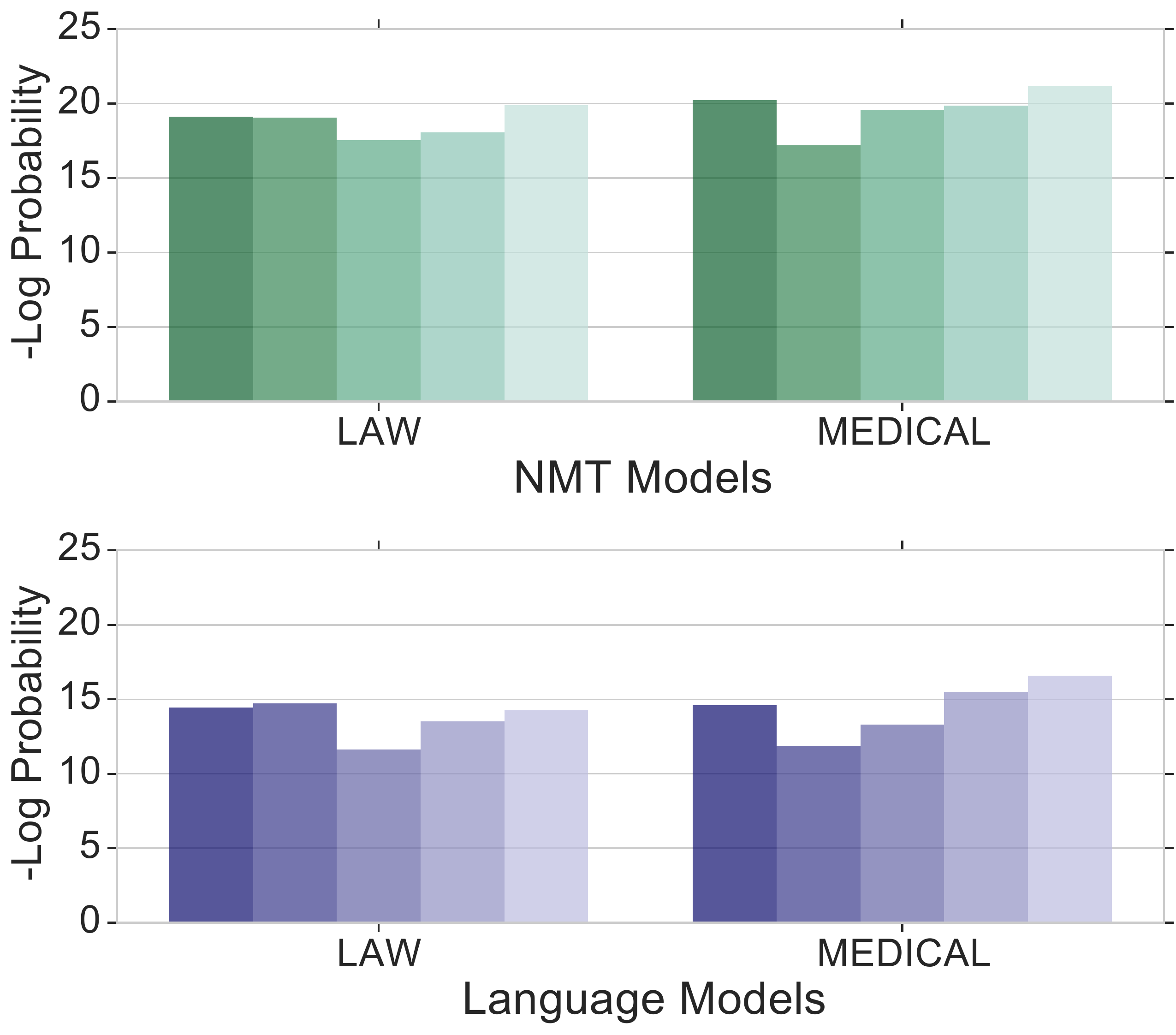}
\caption{ Mean log probabilities of NMT models and LMs trained on law and medical domains for the words ("needle", "hepatic", "complete", "justify", "suspend"). LM and NMT probabilities are correlated for each domain. (More examples in Section~\ref{sec:dom}.)} 
\label{fig:vis}
\end{figure}

\vspace{0pt}

There is a rich literature in domain adaptation for neural networks \cite{luong2015stanford,tan2017distant,chu2017empirical,ying2018transfer}. 
In particular, we focus on two lines of work that are conducive to \emph{unsupervised} adaptation, where there is no training data available in the target domain. 
The first line of work focuses on aligning representations of data from different domains with the goal of improving data sharing across the two domains using techniques such as mean maximum discrepancy \cite{long2015learning} or adversarial training \cite{domain_adv_nn,sankaranarayanan2017generate}.
However, these methods attempt to smooth over the differences in the domains by learning domain-invariant features, and in the case when these differences are actually necessary for correctly predicting the output, this can lead to sub-optimal performance \cite{xie2017controllable}.
Another line of research tries to directly integrate in-domain models of other tasks to adapt models in the target task.
For example, ~\newcite{gulcehre2015using} use pre-trained in-domain LMs with out-of-domain NMT models by directly using a weighted sum of probability distributions from both models, or fusing the hidden states of both models and fine-tuning.
These LMs can potentially capture features of the in-domain data, but models of different tasks are inherently different and thus coming up with an optimal method for combining them is non-trivial. 
 The main intuition behind our method is that models with different data requirements, namely LMs and NMT models, exhibit similar behavior when trained on the same domain, but there is little correlation between models trained on data from different domains (as demonstrated empirically in Figure~\ref{fig:vis}).
 Because of this, directly adapting an out-of-domain NMT model by integrating an in-domain LM~(i.e. with methods in \citet{gulcehre2015using}) may be sub-optimal, as the in-domain and out-of-domain NMT may not be highly correlated.
 However, the \emph{difference} between LMs from two different domains will likely be similar to the \emph{difference} between the NMT models. Based on these observations, we propose a new unsupervised adaptation framework, \dd~(\ddab), that utilizes models of a related task to capture \emph{domain differences}. Specifically, we use LMs trained with in-domain and out-of-domain data, which gives us hints about how to compensate for domain differences and adapt an NMT model trained on out-of-domain parallel data. Although we mainly examine NMT in this paper, the general idea can be applied to other tasks as well.

We evaluate \ddab~in two different unsupervised domain adaptation settings on four language pairs. \ddab~demonstrates consistent improvements of up to 4 BLEU points over an unadapted NMT baseline, and up to 2 BLEU over an NMT baseline adapted using existing methods. An analysis reveals that \ddab~significantly improves the NMT model's ability to generate words more frequently seen in in-domain data, indicating that \ddab~is a promising approach to domain adaptation of NMT and neural models in general. 

\section{Background}
\label{sec:background}

\subsection{Neural Language Models}

Given a sequence of tokens $\mathbf{y} = (y_1, y_2, \cdots,  y_N)$, LMs compute a probability of the sequence $p(\mathbf{y})$ by decomposing it into the probability of each token $y_t$ given the history $(y_1, y_2, \cdots, y_{t-1}) $. Formally, the probability of the sequence $\mathbf{y}$ is calculated as:
$$
p(\mathbf{y}) = \prod_{t=1}^N p(y_t| y_1, y_2, \cdots, y_{t-1}).
$$

LMs are comonly modeled using some variety of recurrent neural networks (RNN; \cite{hochreiter1997long,cho2014learning}), where at each timestep $t$, the network first outputs a context-dependent representation $s_t^{LM}$, which is then used to compute the conditional distribution $p(y_t | y_{<t})$ using a softmax layer.
During training, gradient descent is used to maximize the log-likelihood of the monolingual corpus $Y$:
$$
\max_{\theta_{LM}} \sum_{\mathbf{y}^{i} \in Y } \log p(\mathbf{y}^{i} ; \theta_{LM}) .
$$

\subsection{Neural Machine Translation Models}

Current neural machine translation models are generally implemented in the encoder-decoder framework~\cite{seq2seq,cho2014learning}, where the encoder generates a context vector for each source sentence $\mathbf{x}$ and the decoder then outputs the translation $\mathbf{y}$, one target word at a time.

Similarly to LMs, NMT models would also generate hidden representation $s_t^{NMT}$ at each timestep $t$, and then compute the conditional distribution $p(y_t | y_{<t}, \mathbf{x})$ using a softmax layer. Both encoder and decoder are jointly trained to maximize the log-likelihood of the parallel training corpus $(X, Y)$:
$$
\max_{\theta_{NMT}} \sum_{(\mathbf{x}^{i}, \mathbf{y}^{i}) \in (X, Y) } \log p(\mathbf{y}^{i} | \mathbf{x}^{i}; \theta_{NMT}) .
$$

During decoding, NMT models generate words one by one. Specifically, at each time step $t$, the NMT model calculates the probability of next word $p_{\text{NMT}} (y_t | y_{< t}, \mathbf{x} )$ for each of all previous hypotheses $\{y^{(i)}_{\leq t-1}\}$. After appending the new word to the previous hypothesis, new scores would be calculated and top $K$ ones are selected as new hypotheses $\{y^{(i)}_{\leq t}\}$.

\section{\dd}
\label{sec:methods}
In this section, we propose two  approaches under the overall umbrella of the \ddab~framework: Shallow Adaptation~(DDA-Shallow) and Deep Adaptation~(DDA-Deep). At a high level, both methods capture the domain difference by two LMs, trained on in-domain~(LM-in) and out-of-domain~(LM-out) monolingual data respectively. Without access to in-domain parallel data, we want to adapt the NMT model trained on out-of-domain parallel data~(NMT-out) to approximate the NMT model trained on in-domain parallel data~(NMT-in).  

In the following sections, we assume that LM-in, LM-out as well as the NMT-out model have been pretrained separately before being integrated. 

\subsection{Shallow Adaptation}
Given LM-in, LM-out, and NMT-out, our first method, {\it i.e. shallow adaptation}~(DDA-Shallow), combines the three models only at \textit{decoding} time. 
As we have stated above, at each time step $t$, NMT-out would generate the probability of the next word $p_{\text{NMT-out}} (y_t | y_{<t} , \mathbf{x})$ for each of all previous hypotheses $\{y^{(i)}_{<t}\}$. Similarly, language models LM-in and LM-out would output probabilities of the next word $p_{\text{LM-in}} (y_t| y_{<t})$ and $p_{\text{LM-out}} (y_t | y_{<t})$, respectively.  

For DDA-Shallow, the candidates proposed by NMT-out are rescored considering scores given by LM-in and LM-out. Specifically, at each decoding timestep $t$, the probability of the next generated word $y_t$, is obtained by an interpolation of log-probabilities from LM-in, LM-out into NMT-out.





Formally, the log probability of $y_t$ is
\begin{equation}
\label{eqn:shallow}
\begin{aligned}
    & \log \left(p (y_t)\right) \propto \log \left(p_{\text{NMT-out}} (y_t | y_{<t} , \mathbf{x} )\right) \\
    &+ \beta \left[ \log \left(p_{\text{LM-in}} (y_t | y_{<t})\right) -  \log \left(p_{\text{LM-out}} (y_t | y_{<t})\right) \right],
    \end{aligned}
\end{equation}
where $\beta$ is a hyper-parameter.%
\footnote{Note that this quantity is simply proportional to the log probability, so it is important to re-normalize the probability after interpolation to ensure $\sum_k p(y_t =k) =1$. }

Intuitively, Equation~\ref{eqn:shallow} encourages the model to generate more words in the target domain as well as reduce the probability of generating words in the source domain.

\subsection{Deep Adaptation}
DDA-Shallow only functions during decoding time so there is almost no learning involved. In addition, hyper-parameter $\beta$ is the same for all words, which limits the model's flexibility. Our second more expressive {\it deep adaptation}~(DDA-Deep) method enables the model to learn how to make predictions based on the hidden states of LM-in, LM-out, and NMT-out. We freeze the parameters of the LMs and only fine-tune the fusion strategy and NMT parameters.

Formally, at each time step $t$, we have three hidden states $\mathrm{s}_{\text{LM-out}}^{(t)}$, $\mathrm{s}_{\text{LM-in}}^{(t)}$, and $\mathrm{s}_{\text{NMT-out}}^{(t)}$. 
We then concatenate them and use a gating strategy to combine the three hidden states:

\begin{equation*}
     s_{\text{concat}}^{(t)} = \left[\mathrm{s}_{\text{LM-out}}^{(t)};
   \mathrm{s}_{\text{LM-in}}^{(t)}; 
   \mathrm{s}_{\text{NMT-out}}^{(t)}
   \right] , \eqno{(2.1)}
\end{equation*}
\begin{equation*}
g_{\text{LM-out}}^{(t)},
   g_{\text{LM-in}}^{(t)}, g_{\text{NMT-out}}^{(t)}= F\left(s_{\text{concat}}^{(t)}\right) , \eqno{(2.2)}
\end{equation*}
\begin{equation*}
    \begin{aligned}
     s_{\text{DA}}^{(t)} &=  g_{\text{LM-out}}^{(t)} \odot \mathrm{s}_{\text{LM-out}}^{(t)} 
   + g_{\text{LM-in}}^{(t)} \odot \mathrm{s}_{\text{LM-in}}^{(t)} \nonumber  \\
   &+g_{\text{NMT-out}}^{(t)} \odot \mathrm{s}_{\text{NMT-out}}^{(t)} .
\end{aligned}
 \eqno{(2.3)}
\end{equation*}
Here $F$ is a linear transformation and $\odot$ stands for elementwise multiplication. As the three gating values $g$, we use matrices of the same dimension as the hidden states. This design gives the model more flexibility in combining the states.

One potential problem of training with only out-of-domain parallel corpora is that our method cannot learn a reasonable strategy to predict in-domain words, since it would never come across them during training or fine-tuning. In order to solve this problem, we copy some in-domain monolingual data from target side to source side as in~\newcite{currey2017copied} to form pseudo in-domain parallel corpora. The pseudo in-domain data is concatenated with the original dataset when training the models.

   \begin{table*}[t]
  \centering
  \resizebox{0.9\textwidth}{!}{%
  \begin{tabular}{l|c|c|c|c|c|c||c|c}
   \multicolumn{1}{c|}{\multirow{3}{*}{\bf Method} } & \multicolumn{6}{c||}{\bf De-En} &  \multicolumn{1}{c|}{\bf Cs-En} &  \multicolumn{1}{c}{\bf De-En}
   \\
    \cline{2-9}
  &\multicolumn{2}{c|}{LAW} & \multicolumn{2}{c|}{MED} &  \multicolumn{2}{c||}{IT} & \multicolumn{2}{c}{WMT} \\
   \cline{2-9}
  & MED & IT & LAW & IT & LAW & MED & TED & TED\\
  \hline
  \hline
  \multicolumn{6}{l}{\it w/o copying monolingual data} \\
  \hline 
   ~\newcite{koehn2017six} & 12.1 & 3.5 & 3.9 & 2.0 & 1.9 & 6.5 & - & - \\
   \hdashline
   Baseline & 13.60 & 4.34 & 4.57 & 3.29 & 4.30 & 8.56 & 24.25 & 24.00\\
     \hdashline
 LM-Shallow &  13.74 & 4.41 & 4.54 & 3.41 & 4.29 & 8.15 & 24.29 & 24.03\\
   DDA-Shallow & 16.39* & 5.49* & 5.89* & 4.51* & 5.87* & 10.29* & 26.52* & 25.53*\\
  \hline
  \hline
  \multicolumn{6}{l}{\it w/ copying monolingual data} \\
  \hline 
  Baseline &  17.14 & 6.14 & 5.09 & 4.59 & 5.09 & 10.65 & 25.60 & 24.54\\
  \hdashline
  LM-Deep & 17.74 & 6.01 & 5.16 & 4.87 & 5.01 & 11.88  & 25.98 & 25.12\\
  DDA-Deep & 18.02$^\dag$ & 6.51* & 5.85* & 5.39*  & 5.52$^\dag$  & 12.48* & 26.44* & 25.46$^\dag$  \\
  \hline
  \hline
 \multicolumn{6}{l}{\it w/ back-translated data} \\
  \hline 
  Baseline &  22.89 & 13.36 & 9.96 & 8.03 & 8.68 & 13.71 & 30.12 & 28.88 \\
  \hdashline
  LM-Deep & 23.58 & 14.04 & 10.02 & 9.05 & 8.48 & 15.08 & 30.34 & 28.72 \\
  DDA-Deep & 23.74 & 13.96 & 10.74* & 8.85 & 9.28* & 16.40* & 30.69 & 28.85\\
    \end{tabular}}
    \caption{\label{tab:main2} Translation accuracy (BLEU;~\citet{papineni2002bleu}) under different settings. The first three rows list the language pair, the source domain, and the target domain. ``LAW'', ``MED'' and ``IT'' represent law, medical and IT domains, respectively. We use compare-mt~\cite{neubig19naacl} to perform significance tests~\cite{koehn2004statistical} and statistical significance compared with the best baseline is indicated with $*$ ($p <
0.005$) and $\dag$ ($p < 0.05$).}
  \end{table*}
  
\section{Experiments}
\label{sec:experiment}


\subsection{Setup}

\paragraph{Datasets.}
We test both DDA-Shallow and DDA-Deep in two different data settings. 
In the first setting we use the dataset of \newcite{koehn2017six}, training on the law, medical and IT datasets of the German-English OPUS corpus\footnote{http://opus.nlpl.eu}~\cite{tiedemann2012parallel}.
The standard splits contain 2K development and test sentences in each domain, and about 715K, 1M and 337K training sentences respectively.
In the second setting, we train our models on the WMT-14 datasets\footnote{https://www.statmt.org/wmt14/translation-task.html}~\cite{bojar2014findings} which contain data from several domains and test on the multilingual TED test sets of \citet{duh18multitarget}.\footnote{http://www.cs.jhu.edu/~kevinduh/a/multitarget-tedtalks}
We consider two language pairs for this setting, namely Czech and German to English.
The Czech-English and German-English datasets consist of about 1M and 4.5M sentences respectively and the development and test sets contain about 2K sentences.
Byte-pair encoding~\cite{sennrich2016neural} is employed to process training data into subwords with a vocabulary size of 50K for both settings.

\paragraph{Models.} NMT-out is a 500 dimensional 2-layer attentional LSTM encoder-decoder model~\cite{bahdanau2014neural} implemented on top of OpenNMT~\cite{klein2017opennmt}. LM-in and LM-out are also 2-layer LSTMs with hidden sizes of 500. {Here we mainly test on RNN-based models, but there is nothing architecture-specific in our methods preventing them from being easily adapted to other architectures such as the Transformer model~\cite{Vaswani:2017:NIPS}.} 

\paragraph{Baselines.}
We compare our methods with three baseline models: 1) Shallow fusion and deep fusion~\cite{gulcehre2015using}: they directly combine LM-in with NMT-out\footnote{To ensure the fairness of comparison, we use our gating formula~(Equation (2.2)) and fine-tune all parts of NMT-out for deep fusion.}. Shallow fusion combines LM-in and NMT-out during decoding while deep fusion learns to combine hidden states of LM-in and NMT-out. We denote shallow fusion and deep fusion as ``LM-Shallow'' and ``LM-Deep''. 2) 
The copied monolingual data model ~\cite{currey2017copied} which copies target in-domain monolingual data to the source side to form synthetic in-domain data.
3) Back-translation~\cite{sennrich2016improving} which enriches the training data by generating synthetic in-domain parallel data via a target-to-source NMT model which is trained on a out-of-domain corpus.

\subsection{Main Results}
\subsubsection{Adapting Between Domains}
The first 6 result columns of Table~\ref{tab:main2} show the experimental results on the OPUS dataset.
We can see the LM-Shallow model can only marginally improve and sometimes even harms the performance of baseline models. On the other hand, our proposed DDA-Shallow model can outperform the baseline significantly by over 2 BLEU points.
This reinforces the merit of our main idea of explicitly modeling the \emph{difference} between domains, instead of simply modeling the target domain itself.

Under the setting where additional copied in-domain data is added into the training set, both LM-Deep and DDA-Deep perform better than the baseline model, with DDA-Deep consistently outperforming the LM-Deep method, indicating the presence of an out-of-domain LM is helpful.
We also compare with back-translation, a strong baseline for domain adaptation. We obtain back-translated data via a target-to-source NMT model and concatenate the back-translated data with the original training data to train models.
Again, \ddab~generally brings improvements over the baseline and LM-Deep with back-translated data.

\subsubsection{Adapting from a General Domain to a Specific Domain}
The last two result columns of Table~\ref{tab:main2} show the experimental results in the WMT-TED setting. As we can see, in this data setting our baseline performance is much stronger than the first setting. Similarly to the previous setting, DDA-Shallow can significantly improve the baseline model by over 2 BLEU points.
However, the DDA-Deep model cannot outperform baselines by a large margin, probably because the baseline models are strong when adapting from a general domain to a specific domain and thus additional adaptation strategies can only lead to incremental improvements.


\section{Analysis}
\subsection{Domain Differences between NMT Models and LMs}
\label{sec:dom}
\begin{figure}[h]
\centering
\includegraphics[width=0.4\textwidth]{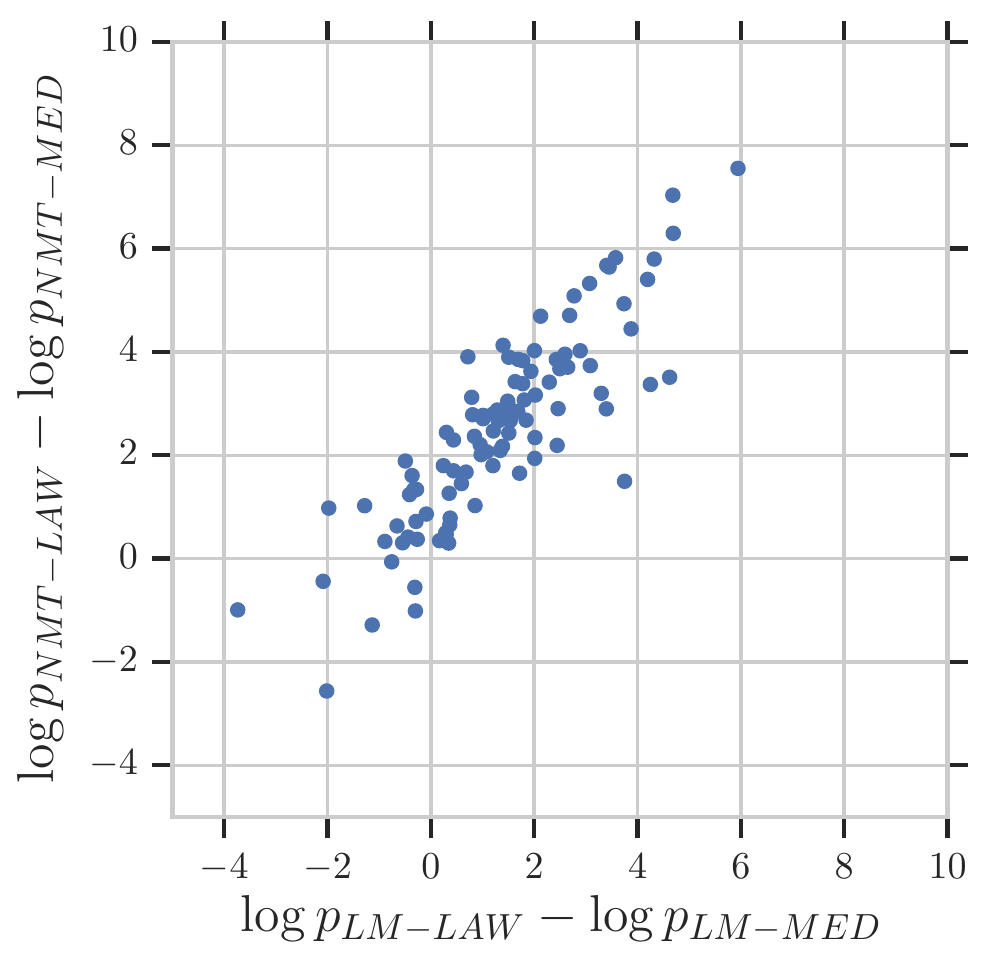}
\caption{Correlation between $\log p_{\text{NMT-LAW}} - \log p_{\text{NMT-MED}}$ and $\log p_{\text{LM-LAW}} - \log p_{\text{LM-MED}}$. We decode each model on the medical set by feeding in the gold labels and calculate the mean of total log probabilities. We plot 100 words that appear frequently in both domains.}
\label{fig:vis_temp}
\end{figure}
In this section, we visualize the correlation between $\log p_{\text{NMT-in}} - \log p_{\text{NMT-out}}$ and $\log p_{\text{LM-in}} - \log p_{\text{LM-out}}$. We treat the law domain as the target domain and the medical domain as the source domain. Specifically, we train four models ${\text{NMT-LAW}}$, ${\text{NMT-MED}}$, ${\text{LM-LAW}}$, ${\text{LM-MED}}$ with law and medical data and decode each model on the medical set by feeding in the gold labels and calculate the mean of total log probabilities, then plot the correlation of 100 words that appear most frequently in both domains. Figure \ref{fig:vis_temp} shows that the difference between NMT models and LMs are roughly correlated, which supports the main motivation of the \ddab~framework. 

 \subsection{Fusing Different Parts of the Models}
 In this section, we try to fuse different parts of LMs and NMT models. Prior works have tried different strategies such as fusing the hidden states of LMs with NMT models \cite{gulcehre2015using} or combining multiple layers of a deep network \cite{peters2018deep}.
 Therefore, it would be interesting to find out which combination of hidden vectors in our DDA-Deep method would be more helpful.
{Specifically, we try to fuse word embeddings, hidden states and output probabilities}.
 
  \begin{table}[h]
  \centering
  \renewcommand{\arraystretch}{1.1}
  \begin{tabular}{c|c|c}
  \bf Components & \bf LAW-MED &  \bf MED-LAW \\
  \hline
  \hline
  Word-Embed  & 17.43 & 5.26 \\
  \hdashline
  Hidden States &18.02 & 5.85\\
  \hdashline
  \multirow{2}{*}{\shortstack[c]{Word-Embed \& \\  Hidden States}} &\multirow{2}{*}{18.00} & \multirow{2}{*}{5.79} \\
   & & \\
    \end{tabular}
    \caption{\label{tab:part} Performance of DDA-Deep when fusing different parts of models on the law and medical datasets.}
  \end{table}
  
  We conduct experiments on the law and medical datasets in OPUS, and experimental results are shown in Table~\ref{tab:part}.
  We find that generally fusing hidden states is better than fusing word embeddings, and fusing hidden states together with word embeddings does not show any improvements over simply fusing hidden states alone. These results indicate that combining the higher-level information captured by the encoder states is more advantageous for domain adaptation. Also, we found that directly using DDA-Deep to fuse {output probabilities} was unstable even after trying several normalization techniques, possibly because of the sensitivity of output probabilities.

  \begin{figure}[t]
\centering
\includegraphics[width=0.49\textwidth]{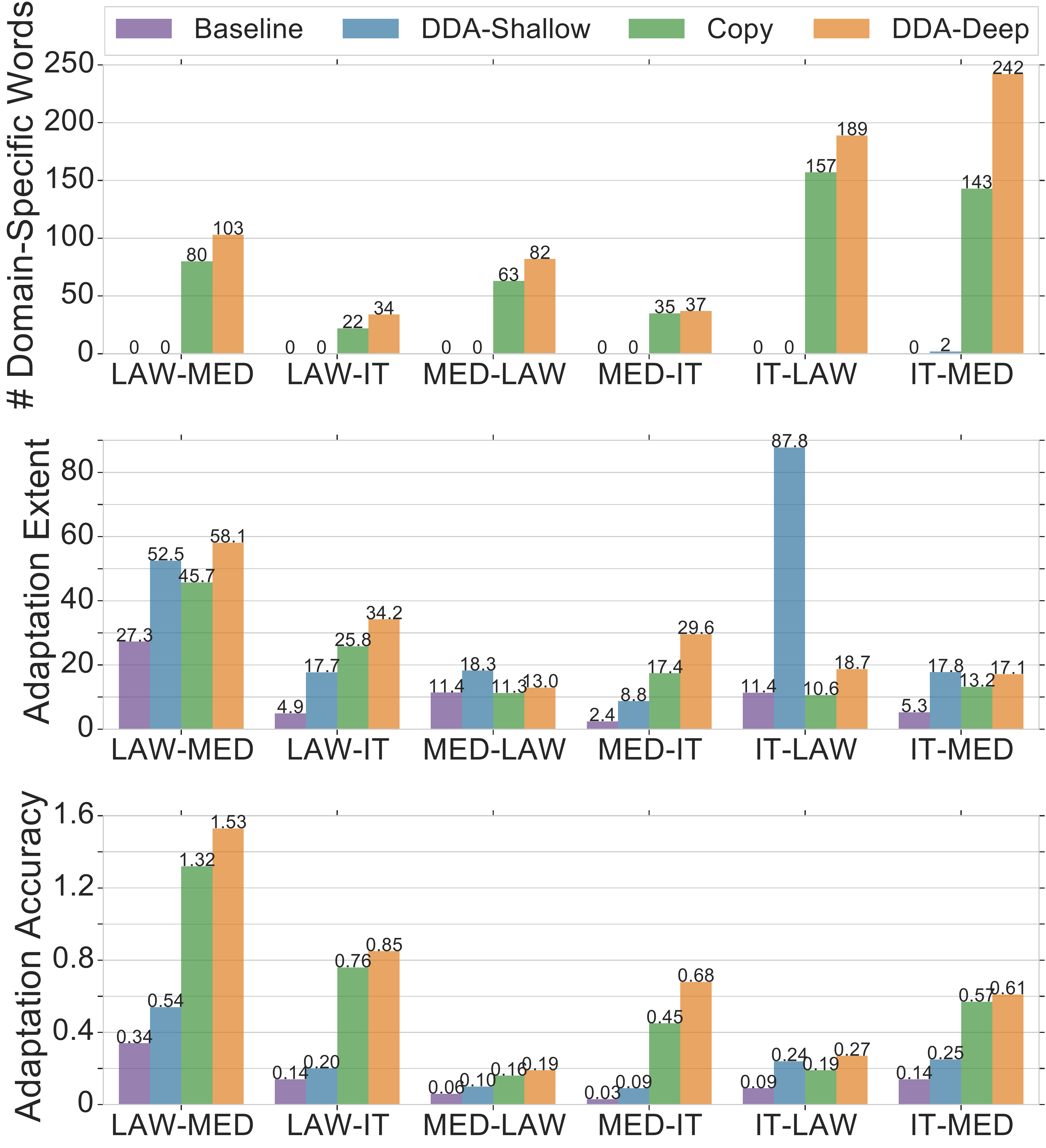}
\caption{\label{fig:bar1}Number of generated domain-specific subwords, scores of adaptation extent and adaptation accuracy for each method. \textit{Top}: count of words only exist in in-domain data produced by different models; \textit{Middle}: adaptation extent of different models; \textit{Bottom}: adaptation accuracy of different models.} 
\end{figure}

\subsection{Analysis of the Adaptation Effect}
\label{adapt_effect}

In this section, we quantitatively and qualitatively analyze the effect of our proposed DDA framework on adapting the NMT model to in-domain data. 
We conduct analysis on the level of the subwords that were used in the MT system, and study whether our methods can generate in-domain subwords that have never appeared or appeared less frequently in the out-of-domain dataset as well as whether our methods can generate these in-domain subwords accurately.

\vspace{2pt}
First, we focus on domain-specific subwords, {\it i.e.} subwords that appear \textit{exclusively} in the in-domain data. The counts of these subwords are shown in Figure \ref{fig:bar1}. In general, both the baseline and DDA-Shallow struggle at generating subwords that never appear in the out-of-domain parallel data. On the other hand, copying monolingual data performs better in this facet, because it exposes the model to subwords that appear only in the in-domain data. DDA-Deep generates the largest number of in-domain subwords among the four models, indicating the effectiveness of our method.

\vspace{3pt}
Second, we propose two subword-level evaluation metrics that study whether the models can generate in-domain subwords and if the generated in-domain subwords are correct. We first define metric ``Adaptation Extent''~(AE) as follows:
\begin{equation*}
\label{eqn:ae}
    \text{AE} = \frac{1}{|V|} \sum_{w \in {V}} \frac{\text{freq\_in}(w)}{\text{freq\_out}(w)} \text{count}(w),\eqno{(3)}
\end{equation*}
where $V$ is the whole vocabulary, $\text{freq\_in}(w)$ and $\text{freq\_out}(w)$ represent the frequency of subword $w$ in both in-domain and out-of-domain corpora, and $\text{count}(w)$ measures how many times subword $w$ appears in the translation result. 

We define ``Adaptation Accuracy'' (AA) in a similar way: 
 \begin{equation*}
 \label{eqn:aa}
     \text{AA} = \frac{1}{|V|} \sum_{w \in {V}} \frac{\text{freq\_in}(w)}{\text{freq\_out}(w)} F1(w),\eqno{(4)}
 \end{equation*}
 where $F1$ denotes the F1-score of subword $w$.
 In order to avoid dividing by zero, we use add-one smoothing when calculating $\text{freq\_out}(w)$. 
 While AE measures the 
 \textit{quantity} of in-domain subwords the models can generate, AA tells us the \textit{quality} of these subwords, namely whether the in-domain subwords form meaningful translations.
 
\vspace{3pt}
We plot the AE and AA scores of our methods as well as the baselines in Figure \ref{fig:bar1}. The AE scores demonstrate that both DDA-Shallow and DDA-Deep adapt the model to a larger extent compared to other baselines even though DDA-Shallow fails to generate domain-specific subwords. 
 In addition, the AA scores reveal that DDA-Deep outperforms other methods in terms of adaptation accuracy while DDA-Shallow is relatively weak in this respect. However, it should be noted that there is still large gap between deep adaptation method and the upper bound where the gold reference is used as a ``translation''; the upper bound is about 10 for each setting.



 \begin{table}[t]
  \centering
  \renewcommand{\arraystretch}{1.2}
    \resizebox{0.5\textwidth}{!}{
  \begin{tabular}{l|l}
 \bf Source & warum wurde Ab- ili- fy zugelassen ? \\ 
 \hline
 \bf Reference & why has Ab- ili- fy been approved ? \\
 \hline
 \hline
 \multirow{2}{*}{\bf Baseline} & {reasons was received why a reminder} \\
  &   was accepted ?\\
 \hdashline
\bf DDA-Shallow & why has been approved? \\
 \hdashline
 \bf Copy & why , \\
 \hdashline
 \bf DDA-Deep & why was Ab- ili- fy authorised ? \\
    \end{tabular}   
    }
    \caption{\label{tab:examples} Translation examples under the law to medical adaptation setting. }
  \end{table}
 
 \vspace{2pt}
 We also we sample some translation results and show them in Table~\ref{tab:examples} to qualitatively demonstrate the differences between the methods. {We could see that by modifying the output probabilities, the DDA-Shallow strategy has the ability to adjust tokens translated by the baseline model to some extent, but it is not capable of generating the domain-specific subwords ``Ab- i li- fy''. However, the DDA-Deep strategy can encourage the model to generate domain-specific tokens and make the translation more correct.} 
 
 \vspace{2pt}
 All of the above quantitative and qualitative results indicate that our strategies indeed help the model adapt from the source to target domains.

  \vspace{2pt}
  \subsection{Necessity of Integrating both LMs}
\label{continued}

    \vspace{2pt}
In this section, we further examine the necessity of integrating both in-domain and out-of-domain LMs. Although previous experimental results partially support the statement, we perform more detailed analysis to ensure the gain in BLEU points is because of the joint contribution of LM-in and LM-out.

\vspace{0pt}
\paragraph{Ensembling LMs.} Ensembling multiple models is a common and broadly effective technique for machine learning, and one possible explanation for our success is that we are simply adding more models into the mix. To this end, we compare DDA-Deep with three models: the first one integrates NMT-out with two LMs-in trained with different random seeds and the second one integrates NMT-out with two LMs-out; we also integrate two general-domain LMs which are trained on both the in-domain and out-of-domain data and compare the performance. The experimental results are shown in Table~\ref{tab:ensemble}.

\vspace{2pt}
We can see that \ddab-Deep achieves the best performance compared with the three other models, demonstrating the gain in BLEU is not simply because of using more models.

\begin{table}[t]
  \centering
  \renewcommand{\arraystretch}{1.1}
  \begin{tabular}{c|c|c}
  \bf Strategy &  \bf LAW-MED & \bf MED-LAW \\
  \hline
  \hline
  LM-in + LM-out  & 18.02 & 6.51 \\
  \hdashline
  two LMs-in  &17.60 & 6.06\\
  \hdashline
  two LMs-out& 17.42 & 6.03  \\
  \hdashline
  two LMs-general & 17.64& 6.22 \\  
    \end{tabular}
    \caption{\label{tab:ensemble} Performance of ensembling different LMs on the law and medical datasets. LMs-general are trained with both in-domain and out-of-domain datasets. }
  \end{table}

\paragraph{Continued Training.} In this section, we attempt to gain more insights about the contribution of LM-in and LM-out by investigating how \ddab-Deep behaves under a continued training setting, where a small number of in-domain parallel sentences are available. We first train the NMT-out model until convergence on the out-of-domain corpus, and then fine-tune it with \ddab-Deep on the in-domain corpus. Here we use the medical and IT datasets as our out-of-domain and in-domain corpora respectively, mainly because the baseline model performs poorly under this setting.
 We randomly select $10,000$ parallel sentences in the in-domain dataset for continued training.
 
 \begin{table*}[ht]
  \centering
  \begin{tabular}{l|c|c|c|c|c|c|c}
  Coverage penalty $\beta$ & 0.00 & 0.05 & 0.10 & 0.15 & 0.20 & 0.25 & 0.30 \\
  \hline
  \hline
  Baseline (no LMs) & 15.61 & 16.28 & 17.26 & \textbf{17.59} & 17.21 & 16.39 & 15.96\\
  \hdashline
  {LM-Deep (LM-out)} & 13.56 & 14.61 & 15.52 & 15.92 & \textbf{15.98} & 15.76 & 15.24 \\
  \hdashline
  LM-Deep (LM-in) & 12.00 & 13.36 & 14.56 & 15.10 & 15.62 & \textbf{15.98} & 15.57 \\
  \hdashline
  DDA-Deep (LM-in + LM-out) & 13.36 & 15.18 & 17.52 & 18.46 &  \textbf{18.62} & 18.03 & 17.17 \\
    \end{tabular}
    \caption{\label{tab:ct} BLEU points of models after continued training on the IT development dataset with different values of coverage penalty $\beta$. }
  \end{table*}
 
\begin{figure}[t]
\centering
\includegraphics[width=0.41\textwidth]{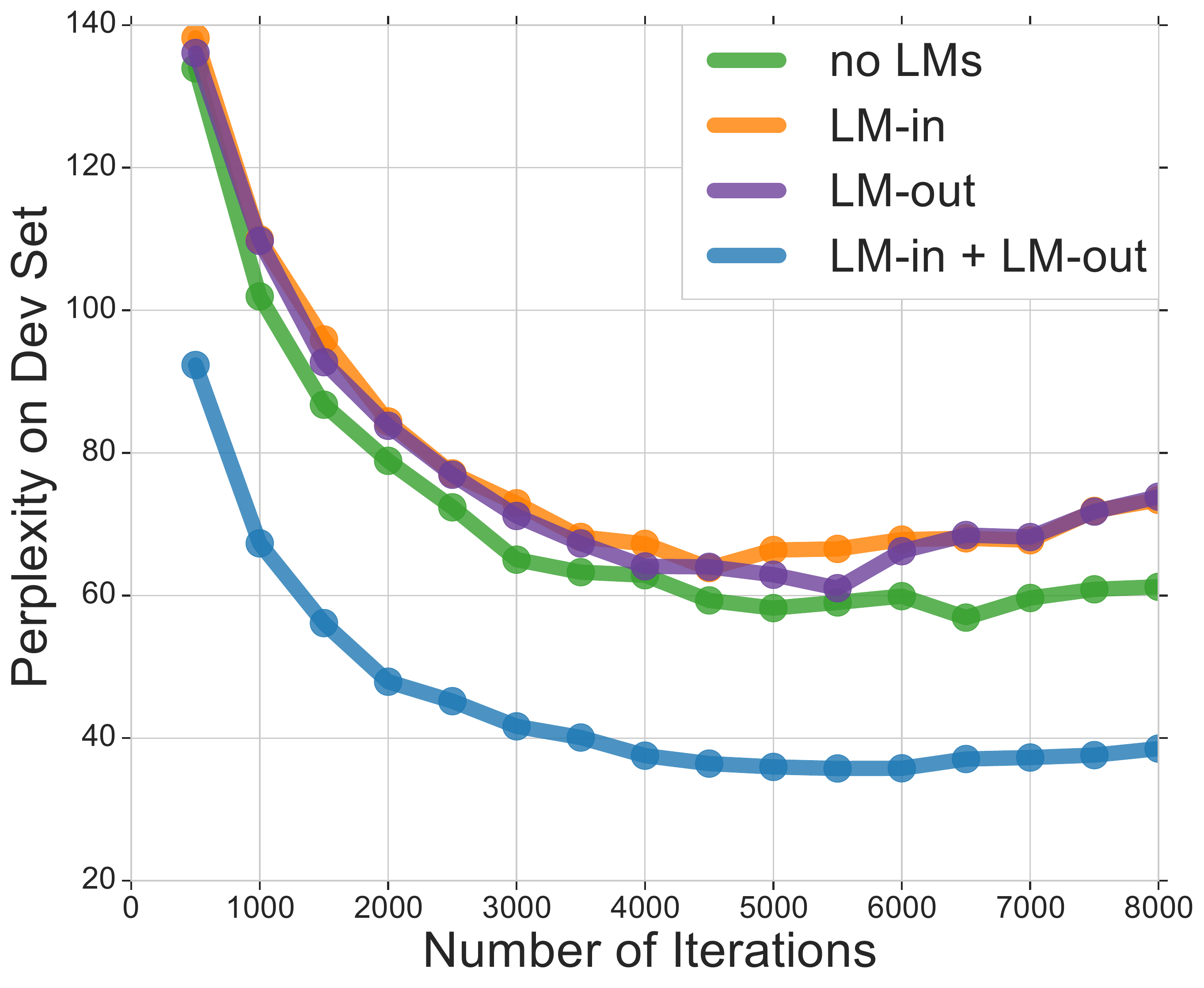}
\caption{ Perplexity on the development set for each method under the continued training setting. ``no LMS'', ``LM-in'', ``LM-out'' and ``LM-in + LM-out'' denote the baseline model, LM-Deep with LM-in, LM-Deep with LM-out and DDA-Deep respectively.}
\label{fig:ct}
\end{figure}

 \vspace{2pt}
 We freeze LM-in and LM-out as before and fine-tune the NMT-out model. The results are shown in Figure~\ref{fig:ct}. We find that the perplexity of deep adaptation method on the development set drops more dramatically compared to baseline models. Figure \ref{fig:ct} shows that integrating only LM-in or LM-out with the NMT model does not help, and sometimes even hurts the performance. 
 This finding indicates that there indeed exists some correlation between LMs trained on different domains. Using both LM-in and LM-out together is essential for the NMT model to utilize the domain difference to adapt more effectively.
 
  \vspace{2pt}
 However, if we look at the BLEU points on the development set, DDA-deep with continued training performs much worse than the baseline model (13.36 vs. 15.61), as shown in Table~\ref{tab:ct} ($\beta = 0$). This sheds light on some limitations of our proposed method, which we will discuss in the next section.

 \vspace{5pt}
\subsection{Limitations of Current {\ddab} Methods}
 \label{sec:lim}
   \vspace{2pt}
 Although our two proposed methods under the \ddab~framework achieve impressive results on unsupervised domain adaptation for NMT, the translation results still fall behind the gold reference by a large margin and the DDA-Deep performs much worse than the baseline model under a continued training setting as demonstrated in previous sections. In this section, we specify some limitations with our proposed methods and list a few future directions.
 
  
 The objectives of LMs and NMT models are inherently different: LMs care more about the fluency whereas NMT models also need to consider translation adequacy, that is, the translations should faithfully reflect the source sentence~\cite{tu2016modeling}. Therefore, directly integrating LMs with NMT models might have a negative impact on adequacy.

  
 To verify this hypothesis, under the continued training setting we adopt a decoding-time coverage penalty~\cite{wu2016google}, which is a simple yet effective strategy to reduce the number of dropped tokens. As shown in Table~\ref{tab:ct}, the coverage penalty can improve the deep adaptation method by more than 5 BLEU points while the baseline model can only be improved by 2 BLEU points. The best DDA-Deep method outperforms the baseline by 1.03 BLEU points.


  
 These results suggest some promising future directions for designing models under the \ddab~framework. Although current {\ddab} methods can extract domain differences from two LMs, they cannot fully reduce the negative effect of LM objective on the NMT model. Therefore, it may be useful to add domain related priors that encourage the in-domain annd out-of-domain LMs to be more distinct, so that they can capture more domain-specific information.
Another possible option is to add extra objectives to LM pretraining so that it can be fused with the NMT model more seamlessly.


  \vspace{2pt}
\section{Related Work}
\label{sec:related}

Finally, we overview related works in the general field of unsupervised domain adaptation, and then list some specific domain adaptation strategies for neural machine translation.

  \vspace{2pt}
\subsection{Unsupervised Domain Adaptation}
Prior unsupervised domain adaptation methods for neural models mainly address the problem by aligning source domain and target domain by minimizing certain distribution statistics. For instance, \newcite{long2015learning} propose deep adaptation networks that minimize a multiple kernel maximum mean discrepancy (MK-MMD) between source and target domains. \newcite{sankaranarayanan2017generate} on the other hand utilize adversarial training to match different domains. 
Researchers have also tried to use language models for unsupervised domain adaptation. For example,~\newcite{siddhant2018unsupervised} propose to apply Embeddings from Language Models (ELMo)~\cite{peters2018deep} and its variants in unsupervised transfer learning. 
  \vspace{2pt}
\subsection{Domain Adaptation for NMT}
Domain adaptation is an active research topic in NMT \cite{chu2018survey}. 
Many previous works focus on the setting where a small amount of in-domain data is available. For instance, continued training~\cite{luong2015stanford,freitag2016fast} is one of the most popular methods, whose basic idea is to first train an NMT model on out-of-domain data and then fine-tune it on the in-domain data.
Also, \newcite{wang2017instance} propose instance weighting methods for NMT domain adaptation problem, the main goal of which is to assign higher weights to in-domain data than out-of-domain data. 

Using LMs or monolingual data to address domain adaptation has been investigated by several researchers~\cite{sennrich2016improving,currey2017copied,hu19acl}. \newcite{moore2010intelligentselection,axelrod2011adaptation} use LMs to score the out-of-domain data and then select data that are similar to in-domain text based on the resulting scores, a paradigm adapted by \newcite{duh2013adaptation} to neural models. ~\newcite{gulcehre2015using} propose two fusion techniques, namely {shallow fusion} and {deep fusion}, to integrate LM and NMT model. Shallow fusion mainly combines LM and NMT model during decoding while deep fusion integrates the two models during training. Researchers have also proposed to perform adaptation for NMT by retrieving sentences or n-grams in the training data similar to the test set~\cite{farajian2017multi,bapna2019non}. However, it can be difficult to find similar parallel sentences in domain adaptation settings. 
  \vspace{2pt}
 
\section{Conclusion}
\label{sec:conclusion}

We propose a novel framework of domain differential adaptation (DDA) that models the differences between domains with the help of models in a related task, based on which we adapt models for the target task. Two simple strategies under the proposed framework for neural machine translation are presented and are demonstrated to achieve good performance. Moreover, we introduce two subword-level evaluation metrics for domain adaptation in machine translation and analyses reveal that our methods can adapt models to a larger extent and with a higher accuracy compared with several alternative adaptation strategies.

However, as shown in our analysis, there are certain limitations for our current methods. Future directions include adding more prior knowledge into our methods as well as considering more sophisticated combining strategies. We will also validate our framework on other pairs of tasks, such as text summarization and language modeling.
\section*{Acknowledgements}
We are grateful to anonymous reviewers for their helpful suggestions and insightful comments. We also thank Junxian He, Austin Matthews, Paul Michel for proofreading the paper.

This material is based upon work generously supported partly by the National Science Foundation under grant 1761548 and the Defense Advanced Research Projects Agency Information Innovation Office (I2O) Low Resource Languages for Emergent Incidents (LORELEI) program under Contract No. HR0011-15-C0114. The views and conclusions contained in this document are those of the authors and should not be interpreted as representing the official policies, either expressed or implied, of the U.S. Government. The U.S. Government is authorized to reproduce and distribute reprints for Government purposes notwithstanding any copyright notation here on.

\bibliography{acl2019}
\bibliographystyle{acl_natbib}

\cleardoublepage

\end{document}